\documentclass[conference]{IEEEtran}
\usepackage{times}

\usepackage[numbers]{natbib}
\usepackage[utf8]{inputenc} 
\usepackage[T1]{fontenc}    
\usepackage{hyperref}       
\usepackage{url}            
\usepackage{booktabs}       
\usepackage{tocloft}
\usepackage{titletoc}
\usepackage{amsfonts}       
\usepackage{nicefrac}       
\usepackage{microtype}      
\usepackage{xcolor}         
\usepackage{amsmath}
\usepackage{multirow}
\usepackage{multicol}
\usepackage{amssymb}
\usepackage[ruled,vlined]{algorithm2e}
\usepackage{colortbl}
\usepackage{adjustbox}
\usepackage{wrapfig}
\usepackage{float}
\usepackage{listings}
\usepackage{color}
\usepackage{subfig}
\usepackage[framemethod=TikZ]{mdframed}
\usepackage{marvosym}
\definecolor{mygray}{gray}{.85}
\definecolor{myhighlight}{RGB}{193,210,240}
\definecolor{verylightgray}{RGB}{240,240,240}
\definecolor{mypurple}{rgb}{0.5137, 0.098, 0.9608}
\definecolor{tablecolor}{HTML}{f5ecfe}

\newcommand{\ccbf}[1]{\cellcolor{tablecolor}$\mathbf{#1}$}
\newcommand{\gray}[1]{\textcolor{gray}{$#1$}}
\newmdenv[  
  backgroundcolor=verylightgray,  
  hidealllines=true,  
  innerleftmargin=8pt,  
  innerrightmargin=8pt,  
  innertopmargin=4pt,  
  innerbottommargin=4pt  
]{graybox}

\begin{document}

\title{R2RGen: Real-to-Real 3D Data Generation for Spatially Generalized Manipulation}

\author{Xiuwei Xu$^{123*}$, Angyuan Ma$^{123*}$, Hankun Li$^{123}$, Bingyao Yu$^{123}$, Zheng Zhu$^{4}$, Jie Zhou$^{123}$, Jiwen Lu$^{123}$\textsuperscript{\Letter}\vspace{0.03in}\\

$^1$Department of Automation, Tsinghua University, $^2$Beijing Key Laboratory of Embodied Intelligence Systems\\
$^3$Institute for Embodied Intelligence and Robotics, Tsinghua University, $^4$GigaAI,\quad$^*$Equal contribution\\

\href{https://r2rgen.github.io}{\textcolor{mypurple}{\textbf{demo-generation.github.io}}}\vspace{-0.1in}}


%


\twocolumn[{%
\renewcommand\twocolumn[1][]{#1}%
\maketitle
\begin{center}
    \centering
    \captionsetup{type=figure}
    \vspace{-6mm}
     \includegraphics[width=0.99\textwidth]{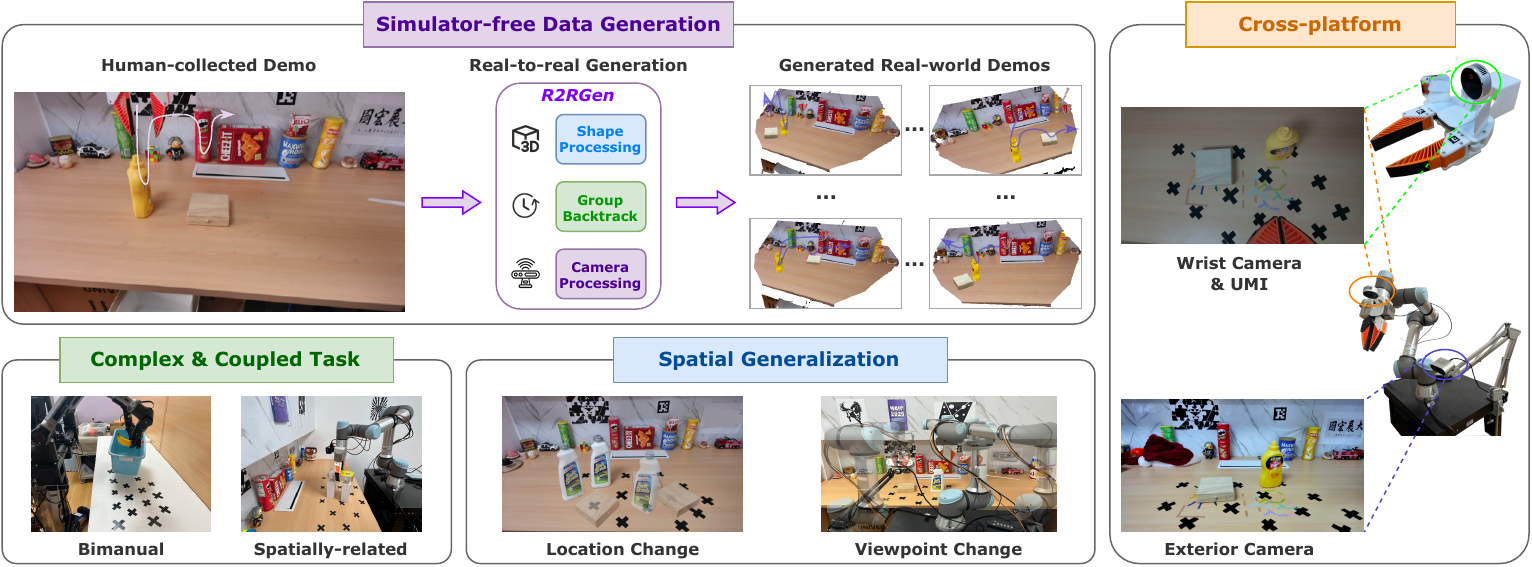}
    \vspace{-1mm}
    \caption{R2RGen is a simulator-free data generation framework. Given one human-collected demonstration, R2RGen directly parses and edits both pointcloud observations and action trajectories in a shared 3D space. R2RGen achieves strong spatial generalization on diverse complex tasks and different data collection setups.}
    \label{fig:teaser}
\end{center}
}]

\begin{abstract}
    Towards the aim of generalized robotic manipulation, spatial generalization is the most fundamental capability that requires the policy to work robustly under different spatial distribution of objects, environment and agent itself. To achieve this, substantial human demonstrations need to be collected to cover different spatial configurations for training a generalized visuomotor policy via imitation learning.
Prior works explore a promising direction that leverages data generation to acquire abundant spatially diverse data from minimal source demonstrations. However, most approaches face significant sim-to-real gap and are often limited to constrained settings, such as fixed-base scenarios and predefined camera viewpoints.
In this paper, we propose a real-to-real 3D data generation framework (R2RGen) that directly augments the pointcloud observation-action pairs to generate real-world data. R2RGen is simulator- and rendering-free, thus being efficient and plug-and-play.
Specifically, we propose a unified three-stage framework, which (1) pre-processes source demonstrations under different camera setups in a shared 3D space with scene / trajectory parsing; (2) augments objects and robot's position with a group-wise backtracking strategy; (3) aligns the distribution of generated data with real-world 3D sensor using camera-aware post-processing.
Empirically, R2RGen substantially enhances data efficiency on extensive experiments and demonstrates strong potential for scaling and application on mobile manipulation.
\end{abstract}

\section{Introduction}

Robotic manipulation with visuomotor policy~\cite{chi2023diffusion,zhao2023learning,fu2024mobile} has achieved great progress in recent years, while the reliance on large amount of human-collected data during imitation learning becomes the main bottlenecks for application and further scaling up~\cite{lin2024data}.
Unlike most prior work focused on fixed-tabletop arms, this paper studies a more general manipulation setting involving mobile manipulators. Since the mobile base may be located at arbitrary positions, the resulting viewpoint variation further increases the policy's reliance on extensive training data~\cite{tan2024manibox}.

\textbf{\emph{Spatial generalization}} is the primary factor driving the substantial data demand during visuomotor policy learning. As pointed out by \cite{garrett2024skillmimicgen}, control difficulty is not uniformly distributed among the human-collected trajectory. Note a trajectory can be divided into two categories of segments: contact-rich segments involving the interaction between robotic arm and objects, and other segments simply indicating the movement of robotic arm in free space, which are also known as \emph{skill} and \emph{motion} segments respectively. Skill segments are generally more challenging, whereas motion segments can often be handled effectively through motion planning.
However, even thought skill segments are more informative, the majority of human demonstration effort is typically devoted to teaching motion behaviors.
For instance, in a task such as “put apple on plate”—even with identical apple, plate, and pick and place skill—hundreds of demonstrations may be needed to cover varying objects' spatial arrangements and robot's base positions to learn a generalized policy.
Therefore, spatial generalization remains a fundamental bottleneck in data efficiency.

To reduce redundant human effort on ensuring spatial generalization, MimicGen~\cite{jiang2024dexmimicgen} and follow-up works~\cite{hoque2024intervengen,garrett2024skillmimicgen,jiang2024dexmimicgen} replace the tedious relocate-and-recollect data collection procedure with automatic demonstration generation. These methods only require a few human-collected data, based on which they augment object configurations and apply transformation and interpolation to generate diverse trajectories with different motion patterns. Though achieving satisfactory performance in sumulation, these methods require on-robot rollouts to collect real-world observation-action pair, which takes much more time and relies on human supervision. 
Recently, DemoGen~\cite{xue2025demogen} introduces a 3D-based data generation method that builds on point-cloud input policy~\cite{ze20243d}. By operating directly in the 3D domain, the approach augments object point clouds to synthesize varied trajectories along with their corresponding visual observations. This pipeline is simulator- and rendering-free, thus being very efficient and avoiding sim-to-real gap. 
However, DemoGen suffers from several critical limitations that restrict its broader applicability: (1) it is restricted to a fixed base and exterior camera, thus unable to handle viewpoint changes or wrist-camera setup; (2) it imposes strong assumption on input data, where the pointcloud of environment should be cropped, limited number of objects are supported, and each skill must involve only one target object; (3) it suffers from visual mismatch problem, i.e., large augmentation leads to incomplete pointcloud observation. 
Due to these constraints, DemoGen does not fully achieve practical real-to-real generation and remains limited in handling diverse task settings.

In this paper, we propose R2RGen, a real-to-real 3D data generation framework which is applicable for mobile robot~\cite{fu2024mobile}, compatible for different camera setups, works on raw pointcloud observation and supports any number of objects and interaction modes.
Given a source demonstration, previous methods apply spatial transformation on each object individually in camera frame. This object-centric paradigm can only handle skills relevant to only one target object and assumes a static camera. To overcome this, we propose to convert pointcloud observations and actions to a shared 3D space and apply group-wise data augmentation, which links each skill to a group of objects rather than a single target to maintain necessary object combination for complicated skills. It also leverages a backtracking mechanism to augment the 3D observation without disturbing the causal order of each operation. Moreover, since pointcloud from 3D sensor (e.g. RGB-D camera) is incomplete, large transformation will make the augmented 3D observation unreasonble. E.g., points should be observed are missing, while points should be occluded exists. To this end, we further present camera-aware 3D post-processing to ensure the 3D observation after augmentation obey the distribution of 3D sensor.
Through extensive evaluation on different camera setups and real-world platforms, we validate R2RGen's effectiveness on one-shot imitation learning and its strong scalability with additional demonstrations, different vision foundation models and task settings.

\section{Related Work}
    \textbf{Imitation learning for robotic manipulation:}
With the development of robotic data collection system and model architecture, using imitation learning to train visuomotor policies~\cite{zhao2023learning,chi2023diffusion,prasad2024consistency,wang2024equivariant,wang2024one,ze20243d,ze2024generalizable}, which end-to-end predict actions from visual observation, becomes a promising way to learn dexterous manipulation skills from human demonstrations.
Inspired by the success of large language models (LLM) and vision language models (VLM), there are multiple recent works exploring scaling up the model size and data amount to train generalist robot policies with imitation learning. One promising approach for training such generalist are vision-language-action models (VLA)~\cite{brohan2023rt,kim2024openvla,black2410pi0,cheang2024gr,zhang2024navid,cheng2024navila,li2024cogact}, which finetunes VLM pre-trained on internet-scale data for robot control.
Though being flexible, imitation learning methods are data-intensive due to the lack of skill priors. In fact, to achieve strong generalization ability, visuomotor policy requires large amount of data for training / finetuning. Since largest embodied datasets~\cite{o2024open,khazatsky2024droid} are still much smaller than the counterparts in vision and language fields~\cite{deng2009imagenet,schuhmann2022laion}, current works manage to solve this problem with advanced data collection systems like UMI~\cite{chi2024universal,ha2024umi} and VR~\cite{cheng2024open,ding2024bunny}, or empirical studies on data scaling~\cite{zhao2024aloha,lin2024data,zha2025guiding} and data selection~\cite{hejna2024re,zhang2025scizor} techniques. 

\textbf{Data generation for visuomotor policy:}
In order to train generalized manipulation visuomotor policy with less human labor, automatic data generation has been paid increased attention in recent years. A branch of works~\cite{hua2024gensim2,wang2023gensim,wang2023robogen,katara2024gen2sim} utilize the common knowledge from LLM / VLM and privileged information from simulatorfor zero-shot task and motion planning. To improve data quality, some works generate robotic data from human demonstration video~\cite{duan2023ar2,lepert2025phantom,yu2025real2render2real}, which fully exploits structured skill primitives from human to acquire reasonable data. However, these methods still rely on vision foundation models~\cite{kim2024garfield,kerr2024robot} to estimate and track poses of hand and object / part, which may not be accurate enough.
Different from generating robotic data in robot-free manner, MimicGen~\cite{mandlekar2023mimicgen} and its follow-up works~\cite{hoque2024intervengen,garrett2024skillmimicgen,jiang2024dexmimicgen} expand real-world demonstrations acquired from teleoperation by synthesizing different execution plans in simulator. These methods work well for simulation, but suffer from time-consuming on-robot rollouts to acquire real-world observation-action pairs.
More recently, DemoGen~\cite{xue2025demogen} proposes to apply augmentation on real-world 3D visual input as well as the trajectory. By using 3D policy, DemoGen directly takes the augmented 3D pointcloud as input, thus being simulator- and rendering-free. 
The real-to-real generation paradigm is efficient and plug-and-play without simulator setup, but currently DemoGen struggles on strong input assumption and visual mismatch problems which severely hinder its application.

\section{Approach}
    \subsection{Problem Statement}\label{sec:statement}
\textbf{Visuomotor policy learning:}
Robotic manipulation task can be modeled as a Partially Observable Markow Decision Process (POMDP) with visuomotor policy $\pi: \mathcal{O}\mapsto\mathcal{A}$, which defines a function that maps current RGB-D observation $o_t\in \mathcal{O}$ to the robot's action $a_t\in\mathcal{A}$. $o_t=(I_t,P_t)$, where $I_t$ is RGB observation and $P_t$ is the pointcloud in camera coordinate system lifted from depth observation and camera intrinsics.
To train this policy, a large dataset of demonstrations $\mathcal{D}=\{o_1^i,a_1^i,...,o_{H_i}^i,a_{H_i}^i\}_{i=1}^N$ should be collected.
To reduce human labor, we aim to improve data efficiency by generating spatially diverse data $\mathcal{D}'$ with only one human-collected source demonstration $D_s\in\mathcal{D}$. Formally written as:
\begin{equation}
    \mathcal{D}'=\{D_s,D_g^1,D_g^2,...,D_g^N\},\ \ \{D_g^i\}_{i=1}^N={\rm R2RGen}(D_s)
\end{equation}
It is expected that we can train a spatially generalized visuomotor policy purely from $\mathcal{D}'$. As a real-to-real framework, we directly augment the 3D observation as well as the action trajectory to generate diverse observation-action pair. Therefore, $\pi$ should be a 3D policy that directly takes in pointcloud as visual input. We opt for iDP3~\cite{ze2024generalizable} as our policy, which consumes the egocentric pointcloud $P_t$ without requirement on camera pose.

\textbf{Assumptions:}
We make the following assumptions: (1) The visuomotor policy only predicts the actions of robotic arm. Although we support manipulation with different base positions, the mobile base remains fixed during each individual task execution. (2) Similar to \cite{mandlekar2023mimicgen}, the actions in a demonstration can be treated as a sequence of continuous end-effector pose and discrete gripper state. Formally, $a_t=(\boldsymbol{{\rm A}}_t^{ee},a_t^{grip})$, where $\boldsymbol{{\rm A}}_t^{ee}$ is SE(3) end-effector pose. 

\textbf{Definition of coordinate system:}
The original end-effector actions are defined in the robot's base coordinate system (BCS), with the arm base $\boldsymbol{{\rm B}}_a$ as its origin. We introduce a world coordinate system (WCS) for the scene and redefine actions $\boldsymbol{\mathrm{A}}_t^{ee}$ by transforming them into this unified frame. The camera frame used for observation is designated as the robot-observation coordinate system (RoCS), which may differ between data collection ($\text{RoCS}_{DC}$) and deployment ($\text{RoCS}_{DP}$).

We support three camera setups: (1) data collection and deployment using an exterior camera which is fixed along with the robot's base; (2) data collection with an exterior camera, while deploying using a wrist camera mounted on the end-effector; (3) data collection and deployment using a wrist camera.
Note that the SE(3) end-effector action $\boldsymbol{\mathrm{A}}_t^{ee}$ naturally defines the transformation from WCS to the wrist camera frame, since the wrist camera is rigidly attached to the end-effector.

\textbf{Overview of R2RGen:}
R2RGen employs a three-stage data generation pipeline to extend the source demonstration:
\begin{equation}
    \text{RoCS}_{DC}\stackrel{SP}{\longrightarrow}\text{WCS}\stackrel{Aug}{\longrightarrow}\text{WCS}\stackrel{CAP}{\longrightarrow}\text{RoCS}_{DP}
\end{equation}
where the source demonstration is first converted from $\text{RoCS}_{DC}$ to WCS with source pre-processing (SP). Then pointcloud observations and actions are jointly augmented (Aug) in WCS. Finally R2RGen applies camera-aware processing (CAP) to convert the pointcloud observation back to $\text{RoCS}_{DP}$.

\subsection{Source Demonstration Pre-processing}

Given the source demonstration $D_s=\{o_1,a_1,...,o_{H_s},a_{H_s}\}$, we need to fully parse the 3D observations $\{P_t\}$ into editable composition of objects, and parse the action trajectory $\{a_t\}$ into motion and skill segments, which facilitates further data generation. Formally, the goal of this pre-processing stage is: 
(1) \textbf{Scene parsing}. Assume the task of $D_s$ involves $K$ relevant objects. So for each timestamp $t$, the pointcloud observation $o_t$ should be segmented into $K$ object pointclouds $\{P_t^1,...,P_t^K\}$, one environment pointcloud $P_t^e$ and the pointcloud $P_t^a$ of robot's arm.
(2) \textbf{Trajectory parsing}. Same with previous definition~\cite{garrett2024skillmimicgen,xue2025demogen} of motion and skill, $\{a_t\}$ should be segmented into sequence of interleaved motion / skill segments $\{a_1,...,a_{m_1}\}$ (motion-1), $\{a_{m_1+1},...,a_{s_1}\}$ (skill-1), $\{a_{s_1+1},...,a_{m_2}\}$ (motion-2), $\{a_{m_2+1},...,a_{s_2}\}$ (skill-2), etc. The objects relevant to each skill should also be known. 

\begin{figure}
    \centering
    \includegraphics[width=1.0\linewidth]{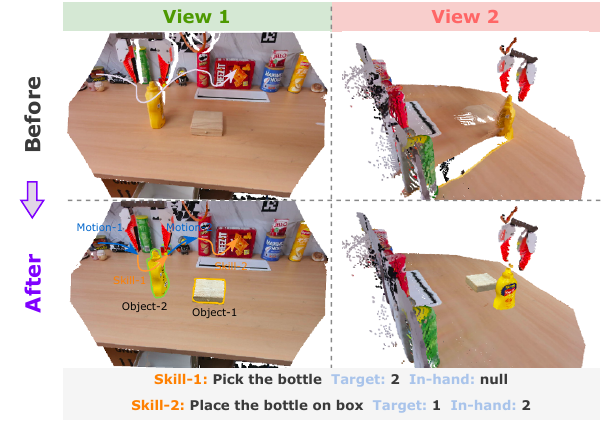}
    \caption{Pre-processing results. The 3D scene is parsed into complete objects, environment and robot's arm. The trajectory is parsed into interleaved motion and skill segments.}
    \label{fig:demo_pre}
\end{figure}

\begin{figure*}
    \centering
    \includegraphics[width=1.0\linewidth]{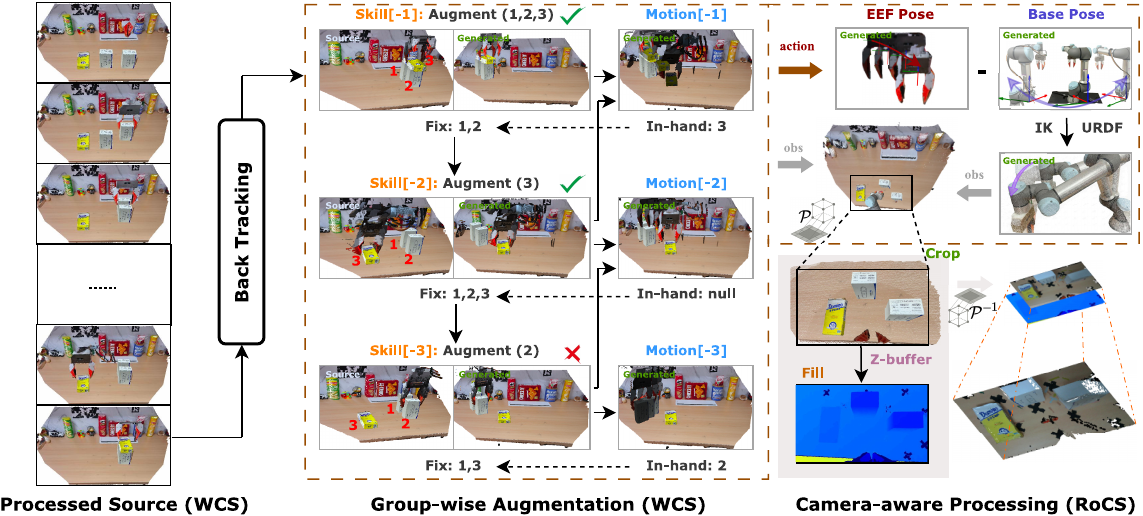}
    \caption{The pipeline of \emph{R2RGen}. Given processed source demonstration, we backtrack skills and apply group-wise augmentation to maintain the spatial relationships among target objects, where a fixed object set is maintained to judge whether the augmentation is applicable. Then motion planning is performed to generate trajectories that connect adjacent skills. After augmentation, we perform camera-aware processing to make the pointclouds follow distribution of RGB-D camera. The solid arrows indicate the processing flow, while the dashed arrows indicate the updating of fixed object set.}
    \label{fig:augment_pipeline}
\end{figure*}

\textbf{Scene parsing:} 
We can segment the $K$ objects in the first frame $I_1$ and track through the whole video. The 2D masks can be projected into 3D to obtain pointcloud of each object. However, only segmenting each objects from $\{P_t\}$ is insufficient due to the incompleteness of RGB-D observation. For instance, the pointcloud of a cup may only represent the side facing the camera, leaving the occluded side unobserved. As a result, the observation will be incomplete when generating demonstrations from a novel viewpoint relative to the object.
To this end, in addition to segmentation, we further complete the object pointclouds $\widetilde{P}_t^i=\mathcal{C}(P_t^i)$. We adopt a template-based 3D object tracking system\footnote{We discuss utilization of other vision foundation models for tracking and completion in Section \ref{sec:ext} as an extension.}~\cite{wen2024foundationpose} to achieve this, which generates complete object pointclouds $\{\widetilde{P}_t^1,...,\widetilde{P}_t^K\}_{t=1}^{H_s}$ for all frames given $K$ object masks in the first frame $I_1$.
Note that the object pointclouds should be converted to WCS. If the data is collected with an exterior camera, we simply use a fixed transformation $\boldsymbol{{\rm T}}_e$ from exterior camera to WCS. When a wrist camera is used, we apply $(\boldsymbol{{\rm A}}_t^{ee})^{-1}$ to convert the pointcloud of corresponding timestamp from wrist frame to WCS.

We also need to acquire the complete environment pointcloud $\widetilde{P}^e$. Since $\widetilde{P}^e$ is static regardless of time, we can simply remove the $K$ objects and scan the environment in WCS before we collect $D_s$.
The pointcloud of robot's arm is ignored at this stage. We just measure $\boldsymbol{{\rm B}}_a$ of BCS, including the base position and orientation, which will be used to generate complete arm pointcloud according to robot dynamics in the following stage.

\textbf{Trajectory parsing:}
We introduce a lightweight annotation system. The interface plays the RGB video $\{I_1,...,I_{H_s}\}$ and asks the annotator to label the start frame and end frame of each skill. The intermediate trajectories between two skill segments are classified as motion segments. Apart from annotating skill segments, the annotator also specifies the object IDs (ranging from $1$ to $K$) associated with each skill. 
Specifically, for every skill, both target object IDs and in-hand object ID are provided: the in-hand object refers to the item being held by the gripper during the skill execution (if any), while the target objects denote the entity with which the gripper interacts. IDs may be null, a single object, or multiple objects, depending on the task structure.
The entire process uses only the RGB video as input and requires less than 60 seconds per demonstration, making it efficient and minimally labor-intensive—particularly suitable for settings with very few source demonstrations.

Figure \ref{fig:demo_pre} illustrates the parsed results of a source demonstration. More details about the object parsing and trajectory parsing systems can be found in Appendix \ref{sec:annotate}.

\subsection{Group-wise Data Augmentation}

To synthesize new demonstrations, we randomly augment the location and rotation of objects and environment to acquire new scene configurations, and generate the corresponding action trajectories. Previous pointcloud-based data generation methods~\cite{xue2025demogen} assume only one target object per skill and transform skill segments solely based on that object’s transformation. Motion segments are then generated using a planner to connect adjacent skills into a complete trajectory. However, this approach fails when a skill involves multiple target objects whose spatial relationships must be preserved. For instance, for task "build a bridge" shown in Figure \ref{fig:augment_pipeline}, placing the bridge deck (object-3) requires the two bridge piers (object-1 and object-2) to be positioned at a specific relative distance. Independently augmenting each pier would disrupt this relationship and prevent successful execution of the final skill. To support arbitrary interaction modes, we propose a group-wise augmentation strategy that maintains structural constraints among multiple objects during data generation.

\textbf{Group-wise backtracking:}
Instead of modeling skills as object-centric, we assign each skill to a group of objects consists of the annotated target and in-hand objects. All objects within the same group undergo identical geometric transformations (i.e., the same translation and rotation). Augmentations are performed in a backtracking manner to avoid causal conflicts among object states.
Formally, we begin from the last skill (skill-$n$) $\{a_{m_n+1},...,a_{s_n}\}$. Denote $O_n=O_n^{tar}\cup O_n^{hand}$ is the ID set of target and in-hand objects of current skill, and $\overline{O}_n=\varnothing$ is the ID set of fixed objects (i.e., objects cannot be augmented). We decide whether to augment current group according to $\overline{O}_n\cap O_n$. If the intersection is $\varnothing$, we randomly sample a transformation matrix $\boldsymbol{{\rm T}}_n \in \mathbb{R}^{4\times 4}$ to apply XY-plane translation and Z-axis rotation on group $O_n$ (XY plane is fitted through the tabletop point cloud). Otherwise this group is fixed at current time and we cannot augment it.
After applying the group transformation, the fixed object set is updated as follows:
\begin{equation}\label{eq:fix}
    \overline{O}_{n-1}=(\overline{O}_n \cup O_n^{tar})\setminus O_n^{hand}
\end{equation}
where current group is appended into the fixed set to maintain spatial relationships, while in-hand object is released since its state before grasped is independent of current skill's constraints.
We then proceed to skill-$(n-1)$ and repeat above operations until all skills are traversed.

\textbf{Skill augmentation:}
For skill-$i$, if the corresponding group is not fixed, we apply transformation $\boldsymbol{{\rm T}}_i$ to augment the end-effector's pose while remain the gripper state:
\begin{equation}\label{eq:skill}
    \hat{\boldsymbol{{\rm A}}}_t^{ee}=\boldsymbol{{\rm A}}_t^{ee}\cdot \boldsymbol{{\rm T}}_i,\ \ \hat{a}_t^{grip}=a_t^{grip},\ \ \forall\ t\in[m_i+1,s_i]
\end{equation}
Then the pointclouds of objects $\widetilde{P}_t^k\ (k\in O_i)$ are transformed with $\boldsymbol{{\rm T}}_i$ in the same way.

\textbf{Motion augmentation:}
For motion-$i$, we apply motion planning to generate a trajectory that starts at the end of skill-$(i-1)$ and ends at the beginning of skill-$i$:
\begin{equation*}
    \hat{\boldsymbol{{\rm A}}}_{t_1:t_2}^{ee}=\texttt{MotionPlan}(\hat{\boldsymbol{{\rm A}}}_{s_{i-1}},\hat{\boldsymbol{{\rm A}}}_{m_i+1}),\ \ \hat{a}_{t_1:t_2}^{grip}=a_{t_1:t_2}^{grip},
\end{equation*}
\begin{equation}\label{eq:motion}
    t_1=s_{i-1}+1,\ \ t_2=m_i
\end{equation}
Then the pointclouds of in-hand objects $\widetilde{P}_t^k\ (k\in O_i^{hand})$ are transformed with relative pose transformation $(\boldsymbol{{\rm A}}_t^{ee})^{-1}\cdot \hat{\boldsymbol{{\rm A}}}_t^{ee}$.

\textbf{Base augmentation:}
We augment the base position and orientation $\boldsymbol{{\rm B}}_a$ to $\hat{\boldsymbol{{\rm B}}}_a$ using transformation $\boldsymbol{{\rm T}}_b$, which simulates the viewpoint change of robot. With the augmented end-effector pose and arm base in WCS, the joint angles of robot's arm can be solved using Inverse Kinematics in BCS. In this way, the complete arm pointcloud $\widetilde{P}_t^a$ can be acquired through:
\begin{equation}\label{eq:dyna}
    \widetilde{P}_t^a=\mathcal{U}(\hat{\boldsymbol{{\rm B}}}_a,\texttt{IK-Solver}(\hat{\boldsymbol{{\rm B}}}_a,\hat{\boldsymbol{{\rm A}}}_t^{ee}))
\end{equation}
where $\mathcal{U}$ converts the robotic arm's URDF model into corresponding pointcloud in BCS based on joint angles solved by IK, and then transforms this pointcloud to WCS using $\hat{\boldsymbol{\mathrm{B}}}_a$.

Finally, we obtain the augmented pointcloud $\hat{P}_t$ as the combination of object, environment and arm pointclouds.
An algorithm diagram of the group-wise augmentation pipeline is detailed in Appendix \ref{sec:alg}. For bimanual manipulation, we additionally introduce constraints to ensure the generated demonstrations executable. Refer to Appendix \ref{sec:bimanual} for more details on bimanual tasks.

\subsection{Camera-aware 3D Post-processing}


After training on generated demonstrations $\mathcal{D}'$, the 3D policy is deployed in real-world with RGB-D camera as input sensor. Therefore, the pointcloud observation in $\mathcal{D}'$ should be similar to the raw RGB-D observation. 
We first convert the augmented pointcloud $\hat{P}_t$ in WCS to $\hat{P}_t^{trans}$ in $\text{RoCS}_{DP}$. If the deployed camera is an exterior camera, as this camera is fixed along with base, we apply $(\boldsymbol{{\rm T}}_e)^{-1}\cdot\boldsymbol{{\rm T}}_b$ to transform the coordinates. When a wrist camera is adopted for deployment, we apply $\hat{\boldsymbol{{\rm A}}}_t^{ee}$ to convert $\hat{P}_t$ to the wrist camera frame.

After coordinate transformation, there are still two main differences between raw observation and $\hat{P}_t^{trans}$: (1) $\hat{P}_t^{trans}$ is over-complete. While for a given perspective of RGB-D camera, raw pointcloud converted from depth image is complete only at this viewpoint. (2) Due to the augmentation on robot's base, the spatial distribution of $\hat{P}_t^{trans}$ may be shifted. To solve this, we propose a camera-aware 3D post-processing to adjust the distribution of generated pointcloud observations:
\begin{equation*}
    \hat{P}_t^{adjust}=\mathcal{P}^{-1}(\texttt{Fill}(\texttt{Z-buffer}(\texttt{Crop}(\{(u_i,v_i,d_i)\})))),
\end{equation*}
\begin{equation}
    \{(u_i,v_i,d_i)\}=\mathcal{P}(\hat{P}_t^{trans})
\end{equation}
where $\mathcal{P}$ projects 3D pointcloud to image plane with camera intrinsics. The $\texttt{Crop}$ operation removes pixels $\{(u_i,v_i,d_i)|u_i<0\ or\ u_i\geq W\ or\ v_i<0\ or\ v_i\geq H\}$ which are out of image boundary. $\texttt{Z-buffer}$ processes overlapped pixels and only keep one pixel with smallest depth value, which removes hidden points at current viewpoint. In practice, we notice the density of pointclouds may not be so high, making front surface unable to hide all points behind. Therefore, we propose a patch-wise Z-buffer operator, where each point with small depth value can hide deeper points in a $r$-radius neighborhood on image plane.
Since the environment is augmented, pixels near the image boundary may be empty (i.e., no point is projected to these pixels). So we $\texttt{Fill}$ the empty pixels by either shrinking the image size or expanding the environment pointcloud, which we detail in Section \ref{sec:ablation}.
Finally, after post-processing in the image plane, we project the pixels back to camera coordinate system. The adjusted pointcloud $\hat{P}_t^{adjust}$ well matches the distribution of RGB-D camera and can be directly fed into our 3D policy during training.

\section{Experiment}

In this section, we evaluate R2RGen through extensive real-world experiments, demonstrating its effectiveness on one-shot imitation learning and how it scales up with more source demonstrations. We also conduct comprehensive ablation studies to explore the optimal design choices. Furthermore, we show R2RGen can be extended to facilitate appearance generalization and mobile manipulation.

\begin{table*}[t]
\centering
\caption{Real-world evaluation of \emph{R2RGen} for spatial generalization. Success rate is reported.}
\label{tab:main}
\resizebox{1.0\linewidth}{!}{
\begin{tabular}{l|cccccc|cc|c}
\toprule
& \multicolumn{6}{c|}{Single-Arm Task} & \multicolumn{2}{c|}{Dual-Arm Task} &\multirow{2}{*}{Averaged} \\
& Open-Jar & Place-Bottle & Pot-Food & Hang-Cup & Stack-Brick & Build-Bridge & Grasp-Box & Store-Item & \\
\midrule
$1$ Source & $3.1$ & $3.1$ & $3.1$ & $3.1$ & $3.1$ & $3.1$ &  $4.2$ & $4.2$ & $3.4$ \\
+DemoGen & $18.8$ & $15.6$ & -- & -- & -- & -- & 16.7 & 16.7 & -- \\
+\textbf{\emph{R2RGen}} &  \ccbf{50.0} & \ccbf{50.0} & \ccbf{37.5} & \ccbf{34.4} & \ccbf{43.8} & \ccbf{34.4} & \ccbf{41.7} & \ccbf{33.3} & \ccbf{40.3}\\
\midrule
\textcolor{gray}{$10$ Source} & \gray{56.3} & \gray{34.3} & \gray{9.4} & \gray{15.6} & \gray{9.4} & \gray{9.4} &  \gray{25.0} & \gray{20.8} & \gray{22.5} \\
\textcolor{gray}{$25$ Source} & \gray{78.1} & \gray{53.1} & \gray{21.9} & \gray{43.8} & \gray{40.6} & \gray{28.1} &  \gray{29.2} & \gray{33.3} & \gray{41.0} \\
\textcolor{gray}{$40$ Source} & \gray{87.5} & \gray{68.8} & \gray{28.1} & \gray{43.8} & \gray{50.0} & \gray{43.8} &  \gray{37.5} & \gray{41.7} & \gray{50.2} \\
\bottomrule
\end{tabular}}
\end{table*}

\begin{table*}[t]
\centering
\caption{Real-world evaluation of \emph{R2RGen} on exterior-wrist and wrist-wrist setup. Success rate is reported.}
\label{tab:setting_2_3}
\resizebox{0.92\linewidth}{!}{
\begin{tabular}{l|cccc|cccc}
\toprule
& \multicolumn{4}{c|}{Exterior-Wrist} & \multicolumn{4}{c}{Wrist-Wrist} \\
& Pot-Food & Hang-Cup & Stack-Brick & Build-Bridge & Pot-Food & Hang-Cup & Stack-Brick & Build-Bridge \\
\midrule
$1$ Source & -- & -- & -- & -- & $3.1$ & $3.1$ &  $3.1$ & $3.1$ \\
+\textbf{\emph{R2RGen}} &  \ccbf{50.0} & \ccbf{37.5} & \ccbf{37.5} & \ccbf{31.3} & \ccbf{40.6} & \ccbf{37.5} & \ccbf{37.5} & \ccbf{28.1} \\
\midrule
\textcolor{gray}{$10$ Source} & -- & -- & -- & -- & \gray{9.4} & \gray{9.4} &  \gray{6.3} & \gray{9.4}  \\
\textcolor{gray}{$25$ Source} & -- & -- & -- & -- & \gray{21.9} & \gray{31.3} &  \gray{31.3} & \gray{28.1}  \\
\textcolor{gray}{$40$ Source} & -- & -- & -- & -- & \gray{50.0} & \gray{43.8} &  \gray{37.5} & \gray{41.7}  \\
\bottomrule
\end{tabular}}
\end{table*}

\subsection{Experimental Setup}
\textbf{Policy.} We select iDP3~\cite{ze2024generalizable} as the visuomotor policy, which takes egocentric pointclouds (i.e., pointclouds in camera coordinate system) and proprioception state as inputs without requirement on camera pose and calibration. 
Details on training iDP3 on each task are described in Appendix \ref{sec:hyper}.

\textbf{Hardware.} We utlize two robot platforms: single-arm and bimanual. The single-arm platform consists of a 7-DoF UR5 arm equipped with a parallel jaw gripper and a mobile base. An ORBBEC femto bolt camera can be either rigidly affixed via a mounting bracket to the mobile base as exterior camera, or attached to the end-effector as wrist camera, which provides the RGB-D observations. The bimanual platform adheres to the MobileAloha architecture~\cite{fu2024mobile}, employing dual AgileX PiPER arms integrated with HexFellow omnidirectional mobile base. A head-mounted RGB-D camera (exterior camera) provides egocentric perception. See Appendix \ref{sec:hardware} for further details.

\textbf{Tasks and evaluation.} We design 8 representative tasks for evaluation, including 2 simple tasks (Open-Jar, Place-Bottle), 4 complex tasks (Pot-Food, Hang-Cup, Stack-Brick, Build-Bridge) and 2 bimanual tasks (Grasp-Box, Store-Item). We compare with DemoGen~\cite{xue2025demogen} on the two simple single-arm tasks and two bimanual tasks. While for other tasks involving complex spatial relationships among objects, DemoGen fails to generate reasonable data.
We evaluate different methods on diverse objects' locations and rotations as well as robot's viewpoints, including a portion of out-of-distribution samples not encountered during training.
Refer to Appendix \ref{sec:tasks} for detailed task definition and evaluation protocol. 

\textbf{Camera settings.} We conduct experiments on three camera setups mentioned in Section \ref{sec:statement}, which we denote as exterior-exterior, exterior-wrist and wrist-wrist to specify the camera used during data collection and deployment, respectively. Since exterior-exterior is the mainstream setting in prior works, we compare R2RGen with baseline method on all 8 tasks for this setting. For exterior-wrist and wrist-wrist setting, we conduct experiments on the 4 complex single-arm tasks, where the source demonstration is collected using exterior/wrist camera, while policy is evaluated with wrist camera. By default, all experiments employ the exterior–exterior camera setup unless otherwise specified.

\begin{figure*}
    \centering
    \includegraphics[width=1.0\linewidth]{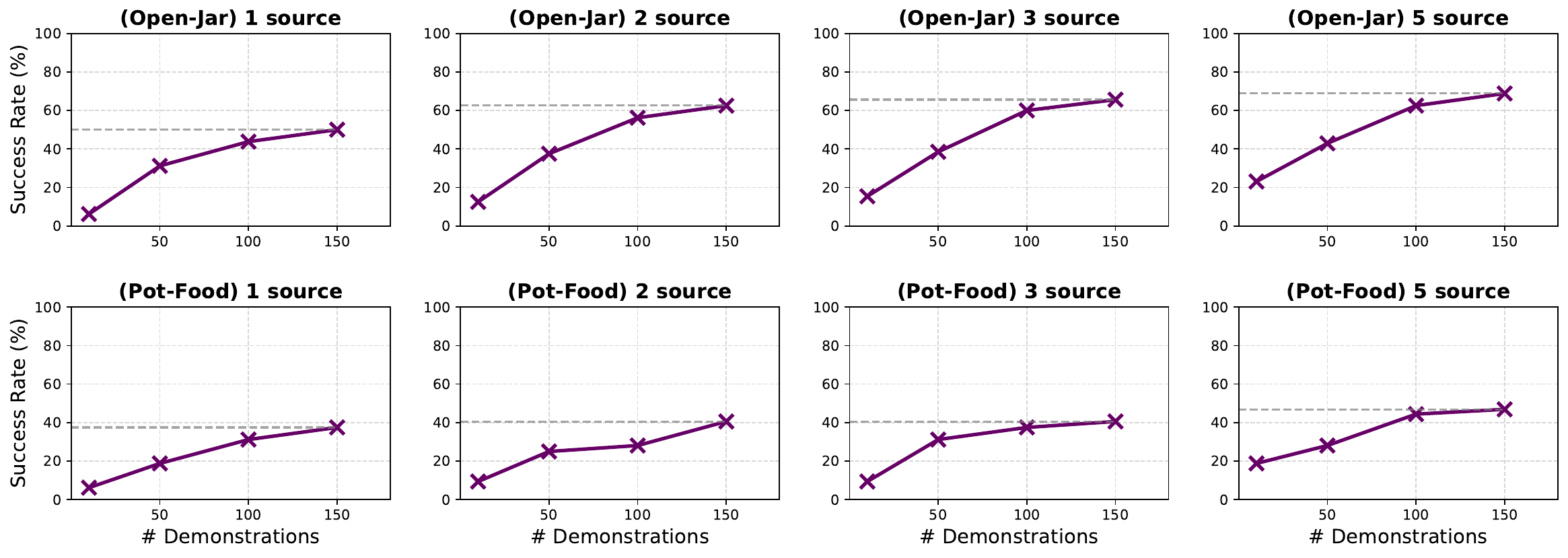}
    \caption{Effects of the number of generated demonstrations and source demonstrations on the final performance of R2RGen.}
    \label{fig:tradeoff}
\end{figure*}

\subsection{Results: One-shot Imitation Learning}
For one-shot imitation learning, we only collect one human demonstration for R2RGen to generate new data. Similar to DemoGen, we replay the collected human demonstration twice to acquire diverse pointcloud observations, which significantly reduce the impact of sensor noise. The three pointcloud trajectories are all used for 3D data generation.
Then we compare the policy trained on purely generated data with ones trained on different number of human demonstrations, as shown in Table \ref{tab:main}.
When trained with only one human demonstration, the policy succeeds merely at the demonstrated pose but fails to generalize. Both DemoGen and R2RGen improve its performance. It is shown that R2RGen consistently outperforms DemoGen across all tasks, even though DemoGen crops pointclouds of background while we do not. This advantage primarily stems from our scene parsing and camera-aware processing techniques, which enable generating high-quality data under large variations in object location / rotation and robot viewpoint. In contrast, DemoGen suffers from significant visual mismatch under such challenging evaluation. R2RGen achieves performance comparable to policies trained with 25 human demonstrations, and even surpasses 40 demonstrations on several difficult tasks, which validates its effectiveness on spatial generalization.

According to Table \ref{tab:setting_2_3}, R2RGen also works well on exterior-wrist and wrist-wrist settings. 
Naive cross-camera deployment—training on exterior views but testing on wrist views—leads to catastrophic failure, underscoring the severe domain gap. By leveraging 3D point cloud transformations between camera and world coordinates, R2RGen eliminates this discrepancy and achieves robust performance gains.

\subsection{Results: Performance-annotation Tradeoff}
We further study how R2RGen scales up with more source demonstrations. For each task, we run R2RGen to generate data from 1, 2, 3 and 5 human demonstrations respectively. We then report the policy performance boost w.r.t. the increase of synthetic data under different numbers of source demonstrations.
As shown in Figure \ref{fig:tradeoff}, the success rate gradually saturates as the number of generated demonstrations increases—a trend also observed with human-collected data. This behavior is due to the limited capacity of the iDP3 policy, which uses a lightweight PointNet encoder. Further scaling beyond this plateau would require policies with larger 3D backbones. We also note that more source demonstrations lead to a higher saturation performance, demonstrating R2RGen’s ability to effectively leverage additional data for improved results.

\begin{table}[t]
\setlength{\tabcolsep}{24pt}
\centering
\caption{Effects of the pointcloud processing.}
\label{tab:ab1}
\begin{tabular}{l|c}
    \toprule
    Method     & SR \\
    \midrule
    Remove pointcloud completion & 34.4 \\
    Remove environment pointcloud & 37.5 \\
    Remove arm pointcloud & 15.6\\
    Remove base augmentation &  21.9 \\
    \textbf{\emph{R2RGen}} & \textbf{50.0} \\
    \bottomrule
\end{tabular}
\end{table}

\begin{table}[t]
\setlength{\tabcolsep}{24pt}
\centering
\caption{Effects of camera-aware processing.}
\label{tab:ab2}
\begin{tabular}{l|c}
    \toprule
    Method     & SR \\
    \midrule
    Remove $\texttt{Crop}$ operation & 34.4 \\
    Remove $\texttt{Z-buffer}$ operation & 15.6 \\
    Remove $\texttt{Fill}$ operation &  28.1\\
    \textbf{\emph{R2RGen}} & \textbf{50.0} \\
    \bottomrule
\end{tabular}
\end{table}

\begin{figure*}
    \centering
    \includegraphics[width=1.0\linewidth]{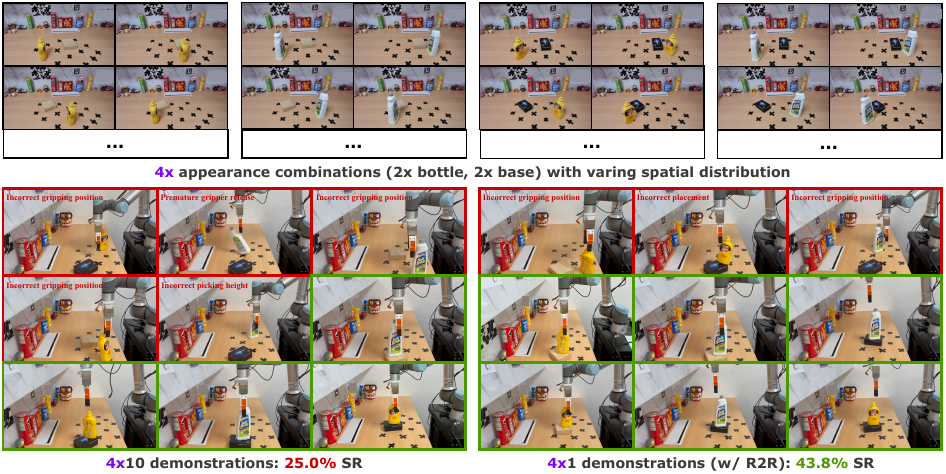}
    \caption{Extension on appearance generalization. The spatial generalization of R2RGen can serve as a foundation to achieve other kinds of generalization with much less data.}
    \label{fig:appearance}
\end{figure*}

\subsection{Ablation Study}\label{sec:ablation}
We carefully design ablation experiments to study the effects of each design choice. The analysis is conducted on the Place-Bottle task.

\textbf{Pointcloud processing.} We first study how object and environment pointclouds affect the final performance, as shown in Table \ref{tab:ab1}. We notice removing pointcloud completion will lead to unrealistic data under large spatial augmentation, while removing operations on environment reduces the policy's robustness to viewpoint changes.

\textbf{Camera-aware processing.} We then ablate camera-aware processing by removing one of the key operations at each time. As shown in Table \ref{tab:ab2}, these operations play a critical role in determining final performance. For the $\texttt{Fill}$ operation, we further compare two design choices, please refer to Appendix \ref{sec:fill} for details.

\subsection{Extension and Application}\label{sec:ext}
\textbf{Extension: non-rigid object.}
Our R2RGen is a general framework that can also handle non-rigid objects. Since currently FoundationPose~\cite{wen2024foundationpose} only supports rigid objects, we can switch to other vision foundation models for non-rigid ones. For instance, ANCSH~\cite{li2020category} for articulated objects and GarmentNets~\cite{chi2021garmentnets} for deformable objects.
More recently, SAM 3D~\cite{chen2025sam} achieves generalized 3D shape generation and pose estimation given monocular RGB observation with depth map, which can perfectly serve as a unified 3D object completion model regardless of rigidness. 
Equipped with this powerful tool, we design two tasks T$_1$ and T$_2$ to validate the universality of R2RGen on articulated objects and deformable objects respectively, as shown in Table \ref{tab:ext_non_rigid}. More details on T$_1$ and T$_2$ can be found in Appendix \ref{sec:aux_task}.

\begin{table}[t]
\centering
\caption{Extension on non-rigid objects using SAM 3D.}
\label{tab:ext_non_rigid}
\resizebox{0.74\linewidth}{!}{
\begin{tabular}{l|cc}
\toprule
& T$_1$-Articulated & T$_2$-Deformable \\
& (Store-Box) & (Cover-Object) \\
\midrule
$1$ Source &  3.1 & 3.1\\
+\textbf{\emph{R2RGen}} &  \ccbf{31.3} & \ccbf{53.1} \\
\midrule
\textcolor{gray}{$10$ Source} & \gray{18.8} & \gray{15.6}  \\
\textcolor{gray}{$25$ Source} & \gray{31.3} & \gray{37.5} \\
\textcolor{gray}{$40$ Source} & \gray{37.5} & \gray{46.9}  \\
\bottomrule
\end{tabular}}
\end{table}


\textbf{Extension: appearance generalization.} 
Beyond spatial generalization, robotic manipulation tasks involve other forms of generalization, such as appearance generalization (i.e., adapting to novel object instances and environments) and task generalization. Among these, spatial generalization serves as the fundamental prerequisite for other generalization capabilities.
Since this work focuses on single-task visuomotor policy learning, we investigate whether the spatial generalization enabled by R2RGen can further facilitate appearance generalization. As shown in Figure~\ref{fig:appearance}, we design a more challenging Place-Bottle task with four distinct bottle-base appearance combinations (2 bottle types $\!\times\!$ 2 base types). We observe that achieving both appearance and spatial generalization significantly increases data demand. Even with 40 human demonstrations (10 per bottle-base pair), the policy only reaches a 25\% success rate.
In contrast, using R2RGen, only 1 demonstration per bottle-base pair (4 in total) is needed to achieve a success rate of 43.8\%, demonstrating its efficiency in handling combined generalization challenges.

\begin{figure}
    \centering
    \includegraphics[width=1.0\linewidth]{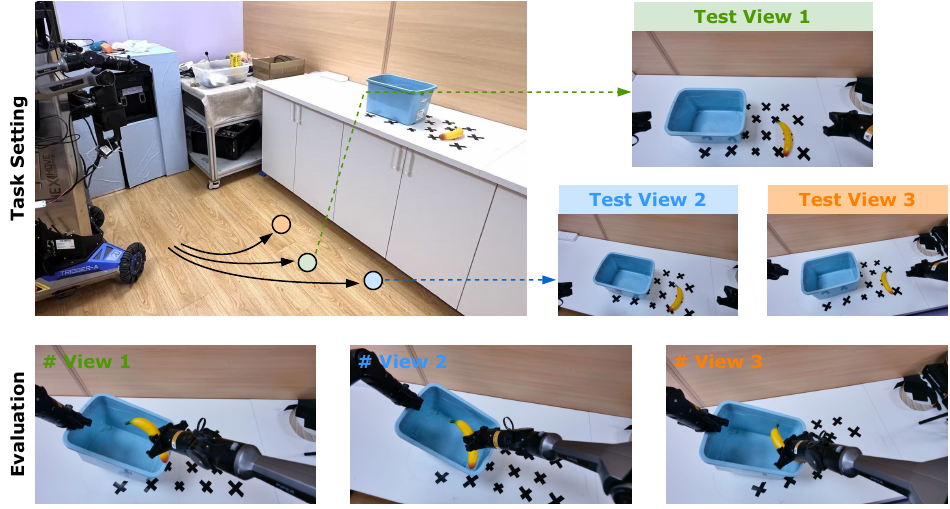}
    \caption{Visualization of mobile manipulation results. The policy trained with R2RGen successfully generalizes to different camera views with only one human-collected demonstration.}
    \label{fig:application}
\end{figure}


\textbf{Application: mobile manipulation.} R2RGen makes our 3D policy achieve strong spatial generalization across different viewpoints without camera calibration, so we can achieve mobile manipulation by simply combining a navigation system~\cite{wu2025moto} and a manipulation policy trained with R2RGen. Since the termination condition of navigation is relatively loose, the robot may stop at different docking point around the manipulation area, which imposes great challenges on the manipulation policy.
According to Figure \ref{fig:application}, using iDP3 trained with R2RGen, the policy successfully generalizes to different docking points with maximum distance larger than 5cm. Different from DemoGen (DP3) which requires a careful calibration of the camera pose to crop environment pointclouds, our method directly applies on raw RGB-D observations during both data generation and policy training / inference stages, which is more practical in real-world applications.

\section{Concluding Remark}
    R2RGen introduces a real-to-real 3D data generation framework that generalizes beyond prior pointcloud-based methods such as DemoGen~\cite{xue2025demogen}. Specifically, it supports mobile manipulators, raw sensor inputs, different camera setups, arbitrary numbers of objects and diverse interaction modes, overcoming key limitations of existing approaches.
With only one human demonstration, our method directly augments the 3D observation as well as the action trajectories to generate large amount of pointcloud-action pairs, which are utilized to train a spatially generalized 3D policy.
Extensive experiments on multiple real-world tasks validate the effectiveness of R2RGen. We further extend its application to different vision foundation models, appearance generalization and mobile manipulation scenarios, demonstrating its strong generalizability and scalability for broad real-world deployment.

\section*{Acknowledgements}
This work was supported in part by the National Natural Science Foundation of China under Grant 624B2076, Grant 62125603, and Grant 62321005, and in part by the Beijing Natural Science Foundation under Grant No. L247009.

\bibliographystyle{plainnat}
\bibliography{references}

\appendix
\subsection{Implementation Details}
\subsubsection{Annotation System}\label{sec:annotate}
There are three stages to collect and process source demonstration, as shown below.

\textbf{Template and environment scanning:} To collect demonstration, we first move the robot to the front of a table or any other platform. Then the robot's base are fixed during task execution. Before we perform teleoperation, we first remove all relevant objects from the table and take a RGB-D image $o_0$, which is converted to WCS as the complete environment $\widetilde{P}^e$.
Then we individually scan each object to acquire their 3D template via the \href{https://developer.apple.com/cn/augmented-reality/tools/}{RealityComposer} App on iPad.

\begin{figure*}
    \centering
    \includegraphics[width=1.0\linewidth]{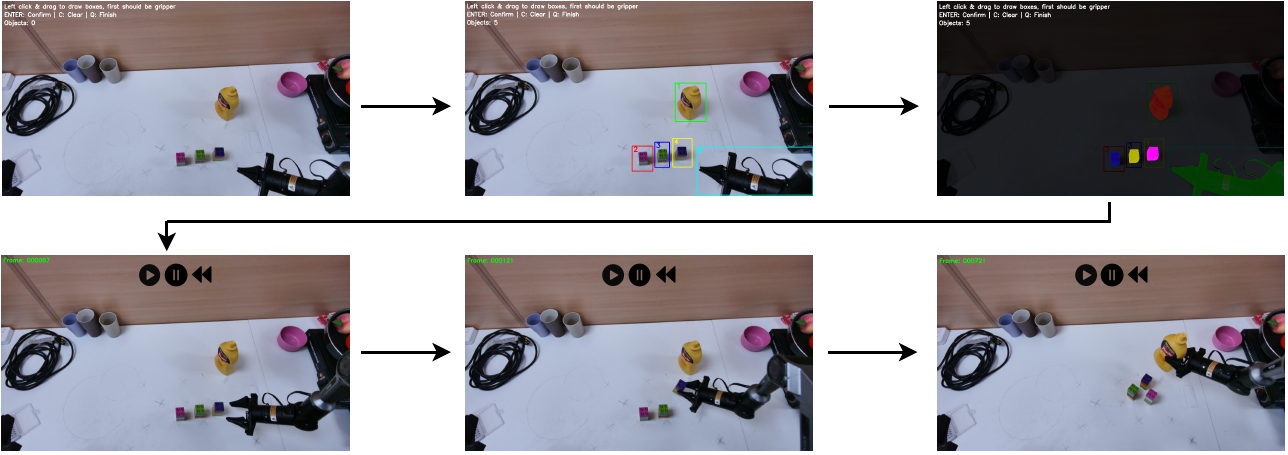}
    \caption{The annotation UI. The users first segment all relevant objects in the first frame. Then they click $\texttt{Play}$, $\texttt{Stop}$ or $\texttt{Rollback}$ to capture key frames for skill / motion annotation.}
    \label{fig:annotate_UI}
\end{figure*}

\textbf{Annotation with UI:} After scanning, we put objects back to the table and start teleoperation to collect sequence of observation-action pair. After that, our UI plays the RGB video and ask user to annotate. As shown in Figure \ref{fig:annotate_UI}. The user first draws boxes on the initial frame to label each object with index, which is then processed with SAM~\cite{kirillov2023segment} to get object masks. Then user watches the video and is able to click $\texttt{Play}$, $\texttt{Stop}$ or $\texttt{Rollback}$ at anytime to capture key frames (i.e., the start / end of skill segments). When the user stops at a key frame, they can press $\texttt{Annotate}$ and enter annotation mode. In this mode, the user is asked to type in the start and end frame of each skill segment, as well as the target and in-hand object IDs corresponding to each skill.
The annotations will be processed into a json file. We show an example as below.

\begin{graybox}
\texttt{\{\newline
\textcolor{verylightgray}{aa}"masks": [\newline
\textcolor{verylightgray}{aaaa}\textcolor{red}{"mask\_gripper.png"},\newline
\textcolor{verylightgray}{aaaa}\textcolor{red}{"mask\_1.png"},\newline
\textcolor{verylightgray}{aaaa}\textcolor{red}{"mask\_2.png"}\newline
\textcolor{verylightgray}{aaaa}\textcolor{red}{"mask\_3.png"}\newline
\textcolor{verylightgray}{aa}],\newline
\textcolor{verylightgray}{aa}"arms": \textcolor{red}{2},\newline
\textcolor{verylightgray}{aa}"annotations": [\newline
\textcolor{verylightgray}{aaaa}\{\newline
\textcolor{verylightgray}{aaaaaa}"frame": \textcolor{red}{4},\newline
\textcolor{verylightgray}{aaaaaa}"type": \textcolor{red}{"motion"}\newline
\textcolor{verylightgray}{aaaa}\},\newline
\textcolor{verylightgray}{aaaa}\{\newline
\textcolor{verylightgray}{aaaaaa}"frame": \textcolor{red}{12},\newline
\textcolor{verylightgray}{aaaaaa}"type": \textcolor{red}{"skill"},\newline
\textcolor{verylightgray}{aaaaaa}"target": \textcolor{red}{[2]},\newline
\textcolor{verylightgray}{aaaaaa}"left\_hand": \textcolor{red}{null},\newline
\textcolor{verylightgray}{aaaaaa}"right\_hand": \textcolor{red}{null}\newline
\textcolor{verylightgray}{aaaa}\}\newline
\textcolor{verylightgray}{aaaa}\{\newline
\textcolor{verylightgray}{aaaaaa}"frame": \textcolor{red}{23},\newline
\textcolor{verylightgray}{aaaaaa}"type": \textcolor{red}{"motion"}\newline
\textcolor{verylightgray}{aaaa}\},\newline
\textcolor{verylightgray}{aaaa}\{\newline
\textcolor{verylightgray}{aaaaaa}"frame": \textcolor{red}{31},\newline
\textcolor{verylightgray}{aaaaaa}"type": \textcolor{red}{"skill"},\newline
\textcolor{verylightgray}{aaaaaa}"target": \textcolor{red}{[1,3]},\newline
\textcolor{verylightgray}{aaaaaa}"left\_hand": \textcolor{red}{[2]},\newline
\textcolor{verylightgray}{aaaaaa}"right\_hand": \textcolor{red}{null}\newline
\textcolor{verylightgray}{aaaa}\}\newline
\textcolor{verylightgray}{aa}]\newline
\}}
\end{graybox}

\textbf{Object tracking and completion:} With the object masks in first frame, the whole RGB-D video and the 3D templates of all objects, we run FoundationPose~\cite{wen2024foundationpose} to track each object across all frames. FoundationPose can accurately predict the 6-DoF pose of each object. So we use the pose to transform object template into WCS to acquire complete object pointclouds. 

\subsubsection{Pipeline of Augmentation}\label{sec:alg}
An algorithm diagram of the group-wise augmentation pipeline is shown in Algorithm \ref{algo}.

\begin{algorithm}
    \caption{Pipeline of Group-wise Augmentation.}\label{algo}
    \KwIn{Trajectory of motion $\{\mathcal{M}_1,...,\mathcal{M}_H\}$ and skill $\{\mathcal{S}_1,...,\mathcal{S}_H\}$; 
    Pointclouds of enironments $\widetilde{P}^e$ and objects $\{\{\widetilde{P}_t^1,...,\widetilde{P}_t^K\}\}$; Robot's base $\boldsymbol{{\rm B}}_a$; 
    Set of target objects $\{O_i^{tar}\}$ and in-hand object $\{O_i^{hand}\}$ of each skill $\mathcal{S}_i$;
    Transformation on each skill $\{\boldsymbol{{\rm T}}_1,...,\boldsymbol{{\rm T}}_H\}$ and base $\boldsymbol{{\rm T}}_b$.}
    \KwOut{Augmented observation $\{\hat{P}_t\}$ and action $\{\hat{a}_t\}$.}
    \BlankLine
    Initialize fixed object set $\overline{O}_H=\varnothing$, backtracking index $T=H$;

    \textcolor{gray}{\tcp{Backtrack skills}}
    \While{$T>0$}{
        \textcolor{gray}{\tcp{If current group is not fixed, augment the trajectory}}
        \If{$(O_T^{tar}\cup O_T^{hand})\cap\overline{O}_T=\varnothing$}{
            Augment $S_T$ with Eq (\ref{eq:skill});
        }
        \textcolor{gray}{\tcp{Otherwise just copy the trajectory}}
        \Else{
            Set $\{\hat{a}_t|t\in \texttt{Timestamp}(S_T)\}=S_T$;
        }
        Update $\overline{O}_T$ with Eq (\ref{eq:fix});

        $T=T-1$;
    }

    \textcolor{gray}{\tcp{Interpolate motions}}
    \While{$T<H$}{
        Interpolate motion $\{\hat{a}_t|t\in \texttt{Timestamp}(\mathcal{M}_{T+1})\}$ with Eq (\ref{eq:motion});

        $T=T+1$;
    }

    \textcolor{gray}{\tcp{Augment 3D observations according to new trajectories}}
    Augment base with $\hat{\boldsymbol{{\rm B}}}_a=\boldsymbol{{\rm B}}_a\cdot\boldsymbol{{\rm T}}_b$;
    
    \While{$T>0$}{
        Acquire arm pointcloud $\widetilde{P}^a_t$ with Eq (\ref{eq:dyna});
        
        Acquire $\hat{P}_t$ with Eq (\ref{eq:aug_pc});

        $T=T-1$;
    }
\end{algorithm}

To augment pointcloud observations, we can divide the pointcloud of whole scene into three parts, i.e., environments, robot's arm and objects:
\begin{equation}\label{eq:aug_pc}
    \hat{P}_t=\widetilde{P}^e\cup\ \widetilde{P}_t^a\cup\texttt{ObjectAugment}_{k=1}^{K}(\widetilde{P}_t^k, \{\hat{a}_t\})
\end{equation}
Here the environment pointcloud is static across all frames. The arm pointcloud is correlated to the pose of end-effector and robot's base. However, different objects are involved in different motion and skill segments, which cannot be augmented as a whole.
Therefore, we traverse all objects. For object $k$, we backtrack all skills that contain this object: $\mathcal{S}_{-1}^k$, $\mathcal{S}_{-2}^k$, ... Then we augment $\widetilde{P}_t^k$ according to the timestamp:
\begin{equation*}
    \hat{P}_t^k=\begin{cases}
        \widetilde{P}_t^k\cdot(\boldsymbol{{\rm A}}_t^{ee})^{-1}\cdot\hat{\boldsymbol{{\rm A}}}_t^{ee}&,\ t\in\texttt{MotionHand}(k)\\
        \widetilde{P}_t^k\cdot\boldsymbol{{\rm T}}_{-1}^k&,\ \text{otherwise}
    \end{cases}\ \ ,
\end{equation*}
\begin{equation*}
    \texttt{End}(\mathcal{S}_{-2}^k)<t
\end{equation*}
\begin{equation*}
    \hat{P}_t^k=\begin{cases}
        \widetilde{P}_t^k\cdot(\boldsymbol{{\rm A}}_t^{ee})^{-1}\cdot\hat{\boldsymbol{{\rm A}}}_t^{ee}&,\ t\in\texttt{MotionHand}(k)\\
        \widetilde{P}_t^k\cdot\boldsymbol{{\rm T}}_{-1}^k\cdot\boldsymbol{{\rm T}}_{-2}^k&,\ \text{otherwise}
    \end{cases}\ \ ,
\end{equation*}
\begin{equation*}
    \texttt{End}(\mathcal{S}_{-3}^k)<t\leq\texttt{End}(\mathcal{S}_{-2}^k)
\end{equation*}
\begin{equation}
    ...
\end{equation}
where $\texttt{End}(\cdot)$ means the end frame of a segment. If the segment does not exist, the value will be set to $-1$. $\texttt{MotionHand}(k)$ represents the set of timestamps that $k$ is in-hand during motion (the in-hand object of motion $\mathcal{M}_i$ equals to that of skill $\mathcal{S}_i$).
During backtracking, every skill involving the target object triggers cumulative application of its spatial transformation to itself and prior timestamps.

\begin{figure}[t]
    \centering
    \includegraphics[width=0.9\linewidth]{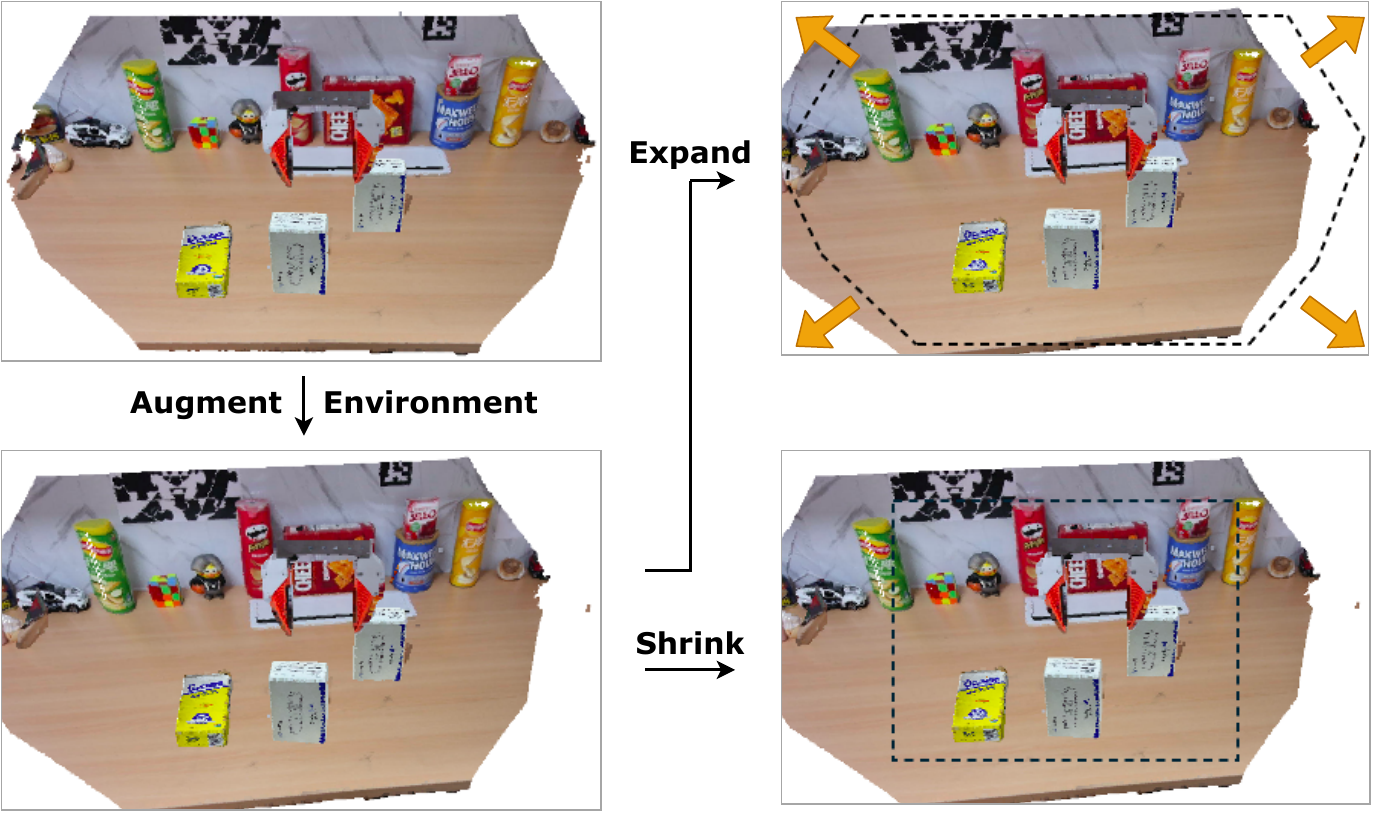}
    \caption{Two implementations of $\texttt{Fill}$ operation, i.e., shrinking and expanding.}
    \label{fig:ab-fill}
\end{figure}

\subsubsection{Bimanual Tasks}\label{sec:bimanual}
As a general framework, R2RGen can also support bimanual manipulation with less modification.
(1) For source pre-processing: we do not segment skill and motion for each arm separately. We just extend the in-hand information to in-left-hand and in-right-hand when annotate each skill. In this way, both single-arm and dual-arm operations can be unified. Then for the arm pointcloud $P_t^a$, we cluster it into two parts to get pointclouds of left arm $P_t^{la}$ and right arm $P_t^{ra}$.
(2) For group-wise augmentation: if in skill $\mathcal{S}_i$, in-left-hand and in-right-hand objects are the same (not null), this means this object is also held by both arms during motion $\mathcal{M}_i$. Therefore, the trajectories of both arms during $\mathcal{M}_i$ should follow a fixed spatial relationship to ensure the object can be stably grasped in both hands. For other cases, we individually interpolate the trajectories for both arm during motion.

\begin{figure*}
    \centering
    \includegraphics[width=0.8\linewidth]{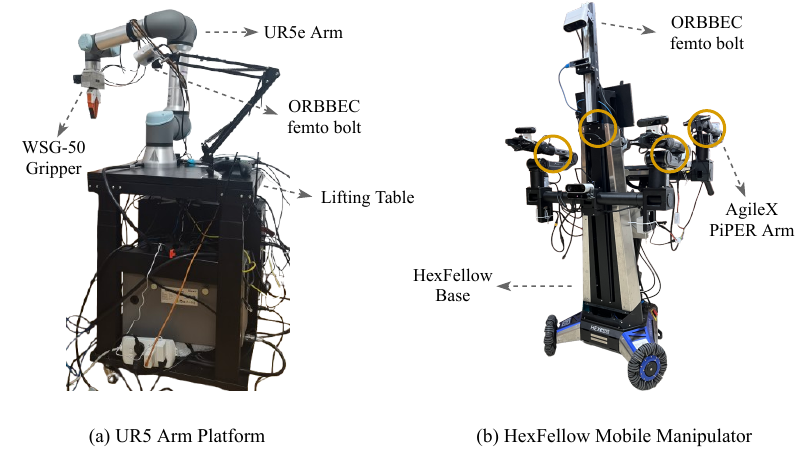}
    \caption{Robot platform overview. We employ two robot platforms: (a) single-arm UR5e system and (b) dual-arm Mobile Aloha system.}
    \label{fig:hardware}
\end{figure*}

\subsubsection{Data Generation and Training}\label{sec:hyper}
Since real-world data always has random noise, one source demonstration may not be enough to cover the distribution of pointclouds. In our experiments, we still collect only one human demonstration, but replay the action trajectory for three times as did in DemoGen. Then we generate demonstrations based on all three source demonstrations.
For each new demonstration generated from a source demonstration, we randomly add small perturbations on the augmented locations and rotations for three times. Specifically, we add random tranlsation within a circle of $1.5$cm radius and random rotation within $\pm20^\circ$.
The total number of generated demonstrations is calculated as $3\!\times\! N\!\times\! 3$, where $N$ is the number of combinations of all augmented locations and rotations.

To train iDP3, denote $T_o$ as the observation horizon, $T_p$ as the action prediction horizon, and $T_a$ as the action execution horizon, we set $T_o=2$, $T_p=16$, and $T_a=8$. The visuomotor policy is run at 5\,Hz. Training was performed for 6,000 epochs on a single RTX 4090 GPU (batch size 64) using Adam (learning rate $1 \times 10^{-4}$, weight decay $1 \times 10^{-6}$). Validation performance plateaued after approximately 2,500 epochs, and we selected the checkpoint with the lowest validation loss.

\subsubsection{Design Choices of Fill Operation}\label{sec:fill}
As shown in Figure \ref{fig:ab-fill}, we ablate two design choices for $\texttt{Fill}$ operation: by shrinking the image size or expanding the environment pointcloud to fill the black dashed outline (i.e., the valid-depth area of the RGB-D camera). If shrinking is adopted, we will apply the same shrinking to raw RGB-D observations when deploying the policy trained with R2RGen.
Empirically, both methods achieve comparable performance; we ultimately adopt shrinking due to its operational simplicity and lack of additional processing requirements.

\begin{figure*}
    \centering
    \includegraphics[width=1.0\linewidth]{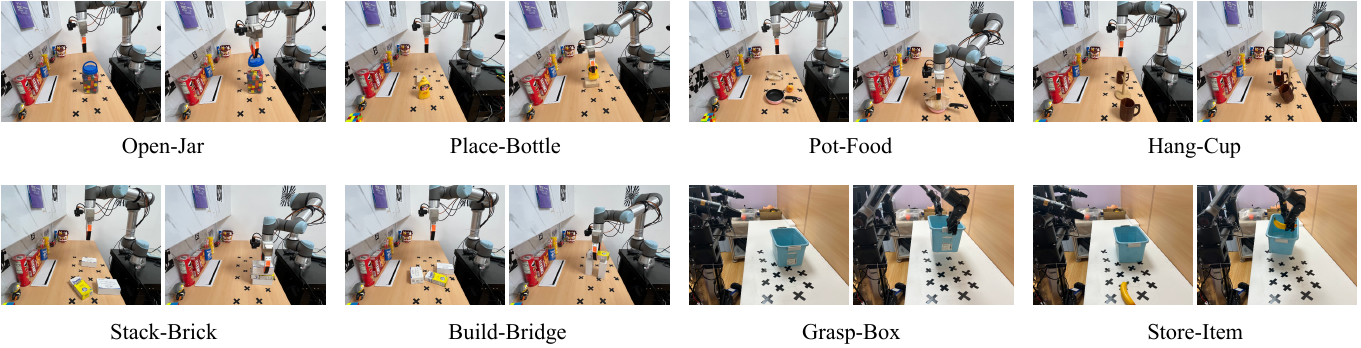}
    \caption{Visualization of our main tasks. We show the start and end moments of each task.}
    \label{fig:tasks}
\end{figure*}

\subsection{Hardware Setup}\label{sec:hardware}

We utilize two robot platforms. The primary platform (Figure \ref{fig:hardware}, a) is a single 7-DoF UR5e arm equipped with a Weiss WSG-50 parallel-jaw gripper. A ORBBEC femto bolt RGB-D camera is mounted to base or wrist to acquire visual observations. The arm is fixed on a height-adjustable table with movable base, which makes it possible for us to evaluate policy with different viewpoint and height. The action space is 7-dimensional (6-DoF end-effector pose plus gripper width). We use a Xbox controller to teleoperate the robotic arm to collect demonstrations.

The second platform (Figure \ref{fig:hardware}, b) follows the design paradigm of Mobile Aloha~\cite{fu2024mobile}, using four AgileX PiPER Arms (two for teleoperation, two for manipulation) and a HexFellow omnidirectional mobile base. We mount one ORBBEC femto bolt RGB-D camera on the robot's head to acquire visual observation. Each robotic arm has 7 dimensions (6-DoF end-effector pose plus gripper width) and the overall action space is 14-dimensional.

\subsection{Tasks and Evaluations}\label{sec:tasks}

\subsubsection{Task Definition}
We carefully design 8 real-world tasks for evaluation, including 6 single-arm tasks and 2 bimanual tasks. A task summary is provided in Table \ref{tab:summary}. We also visualize the tasks in Figure \ref{fig:tasks}.

\begin{table}
\centering
\caption{A summary of our main tasks. \#Obj: number of manipulated objects. \#Eval: number of evaluated configurations. \#Demo: number of generated demonstrations.}\label{tab:summary}
\resizebox{1.0\linewidth}{!}{%
\begin{tabular}{lcccc}
\midrule
Task & Platform &  \#Obj & \#Eval & \#Demo \\
\midrule
\textbf{Open-Jar} & Single-arm & $1$ & $32$ & $144$ \\
\textbf{Place-Bottle} & Single-arm & $2$ & $4 \!\times\! 8$ & $144$ \\
\textbf{Pot-Food} & Single-arm & $3$ & $2 \!\times\! 4 \!\times\! 4$ & $144$ \\
\textbf{Hang-Cup} & Single-arm  & $3$ & $2 \!\times\! 4 \!\times\! 4$ & $144$ \\
\textbf{Stack-Brick} & Single-arm  & $3$ & $2 \!\times\! 4 \!\times\! 4$ & $144$ \\
\textbf{Build-Bridge} & Single-arm  & $3$ & $2 \!\times\! 4 \!\times\! 4$ & $144$ \\
\textbf{Grasp-Box} & Dual-arm & $1$ & $24$ & $108$ \\
\textbf{Store-Item} & Dual-arm  & $2$ & $3\!\times\!8$ & $108$\\
\bottomrule
\end{tabular}}
\end{table}

\begin{figure*}
    \centering
    \includegraphics[width=1.0\linewidth]{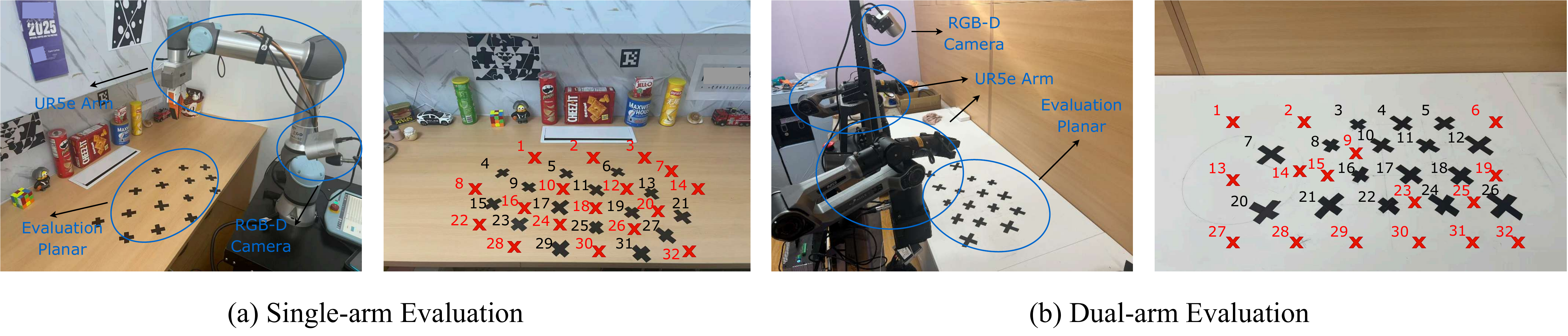}
    \caption{Protocol for evaluating spatial generalization. We evaluate policies on different robot's viewpoints, object locations and rotations. 
    Black crosses indicate seen locations (if human demonstrations are sufficient to cover) and red ones denote unseen locations during training.}
    \label{fig:evaluate_space}
\end{figure*}

We describe these tasks in the text as follows, where \textcolor{orange}{skill} and \textcolor{blue}{motion} verbs are highlighted as orange and blue respectively:

\textbf{(A) Open-Jar.} The gripper \textcolor{blue}{moves} above the jar and \textcolor{blue}{lowers} to an appropriate height. Then it \textcolor{orange}{opens}, \textcolor{orange}{moves down} and \textcolor{orange}{closes} to grasp the handle. It further \textcolor{orange}{rotates} to open the jar.

\textbf{(B) Place-Bottle.} The gripper first \textcolor{blue}{moves} to the bottle and \textcolor{orange}{grasps} it. Then it \textcolor{blue}{lifts} the bottle up and \textcolor{orange}{places} it on the base.

\textbf{(C) Pot-Food.} The gripper first \textcolor{blue}{moves} to the food and \textcolor{orange}{grasps} it. Then it \textcolor{blue}{moves} towards the pot and \textcolor{orange}{puts} food into pot. Next it \textcolor{blue}{moves} to the pot lid and \textcolor{orange}{picks} it up. It finally \textcolor{blue}{moves} towards the pot again and \textcolor{orange}{covers} the pot with the lid.

\textbf{(D) Hang-Cup.} The gripper \textcolor{blue}{moves} to the first cup and \textcolor{orange}{picks} it up. Then it \textcolor{blue}{moves} towards the shelf and \textcolor{orange}{hangs} the cup on the shelf. It repeats the same operation on the second cup, but this time the cup should be \textcolor{orange}{hanged} on a different position of the shelf.

\textbf{(E) Stack-Brick.} The gripper \textcolor{blue}{moves} to the first brick and \textcolor{orange}{picks} it up. Then the gripper \textcolor{blue}{moves} to a designated place and \textcolor{orange}{places} the brick there. It repeats the same operation on the other two bricks to \textcolor{orange}{stack} them one-by-one.

\textbf{(F) Build-Bridge.} The gripper first \textcolor{blue}{moves} to a white box and \textcolor{orange}{picks} it up. Then it \textcolor{blue}{brings} the box to a designated place and \textcolor{orange}{places} it there. The same operation is repeated on the second white box, where the two white boxes (i.e., the bridge piers) should be \textcolor{orange}{placed} in proper distance.
Then the gripper \textcolor{blue}{moves} to and \textcolor{orange}{grasps} the yellow box (i.e., the bridge deck), \textcolor{blue}{moves} towards the bridge piers and \textcolor{orange}{puts} deck on piers to build a bridge.

\textbf{(G) Grasp-Box.} The left gripper \textcolor{blue}{moves} to the left side of the box. Then the right gripper \textcolor{blue}{moves} to the right side. After that, two grippers simultaneously \textcolor{orange}{grasp} the box and \textcolor{orange}{lift} it up.

\textbf{(H) Store-Item.} The left gripper \textcolor{blue}{moves} to the left side of the box. Then it \textcolor{orange}{grasps} the box and \textcolor{orange}{lifts} it up. At the same time, the right gripper \textcolor{blue}{moves} to the banana and \textcolor{orange}{grasps} it. It then \textcolor{blue}{brings} the banana to the box and \textcolor{orange}{stores} the banana into box.

\subsubsection{Evaluation Protocol}

To evaluate spatial generalization, we define large planar evaluation workspaces as illustrated in Figure \ref{fig:evaluate_space}. For each test trial, the initial positions of the objects are determined by sampling distinct locations from a pre-defined set of 32 points on the workspace. Each object is also assigned a random rotation sampled from the range of -20 to 20 degrees, while the robot's base is initialized at one of three distinct locations. 
For tasks involving more than one object, we constrain the range of locations for each object to reduce the number of combinations. We demonstrate the range of object locations of single-arm tasks in Figure \ref{fig:evaluate_space} (a):

\textbf{(A) Open-Jar.} The range of jar is $\{1,2, ..., 32\}$.

\textbf{(B) Place-Bottle.} The range of bottle is $\{4,10,8,17,22,24,28,29\}$, and the range of box is $\{11,13,25,26\}$.

\textbf{(C) Pot-Food.} The range of food is $\{27,31\}$, the range of lid is  $\{6,13,19,21\}$, and the range of pot is $\{9,17,23,29\}$.

\textbf{(D) Hang-Cup.} The range of the first cup is $\{4,16,22,29\}$, the range of the second cup is $\{13,14,20,27\}$, and the range of shelf is $\{11,18\}$.

\textbf{(E) Stack-Brick.} The range of the first brick is $\{8,9,22,23\}$, the range of the second brick is $\{13,19,21,27\}$, and the range of the third brick is $\{10,25\}$.

\textbf{(F) Build-Bridge.} The range of the first bridge pier is $\{8,9,22,23\}$, the range of the second pier is $\{13,14,26,27\}$, and the range of the bridge deck is $\{11,24\}$. 

\begin{figure*}
    \centering
    \includegraphics[width=\linewidth]{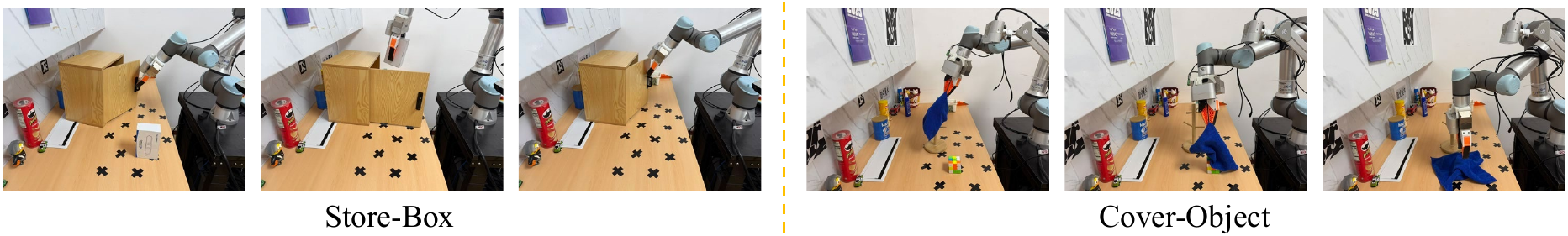}
    \caption{Visualization of our auxiliary tasks involving non-rigid objects. In Store-Box, the box is articulated, while in Cover-Object, the cloth is deformable.}
    \label{fig:tasks_non_rigid}
\end{figure*}

The range of object locations of dual-arm tasks is shown in Figure \ref{fig:evaluate_space} (b):

\textbf{(G) Grasp-Box.} The range of box is $\{1,2,3,4,5,6,7,8,10,11,12,13,20,21,22,23,24,26,27,28,$ $29,30,31,32\}$.

\textbf{(H) Store-Item.} The range of box is $\{1,14,27\}$, and the range of banana is $\{4,6,17,$ $19,23,26,30,32\}$. 

Note the black crosses in Figure \ref{fig:evaluate_space} indicate possibly seen locations during training, where human demonstrations are collected within these locations. The red crosses denote unseen locations which the training set does not cover.

\subsubsection{Auxiliary Tasks}\label{sec:aux_task}
We further design 2 auxiliary tasks, including 1 articulated object task (T$_1$) and 1 deformable object tasks (T$_2$) to validate the effectiveness of R2RGen for handling non-rigid objects. Visualization of the 2 tasks is shown in Figure \ref{fig:tasks_non_rigid}.

\textbf{(T$_1$) Store-Box.} The gripper \textcolor{blue}{moves} to the cabinet handle and \textcolor{orange}{opens} the door. Then the gripper \textcolor{blue}{moves} to the box and \textcolor{orange}{picks} it up. It \textcolor{blue}{moves} to the inside of the cabinet and \textcolor{orange}{places} the box there, and finally \textcolor{orange}{closes} the door. 

\textbf{(T$_2$) Cover-Object.} The gripper \textcolor{blue}{moves} to the shelf and \textcolor{orange}{picks} up the cloth. Then the gripper \textcolor{blue}{moves} to the target object and \textcolor{orange}{places} the cloth over it to cover it.

\end{document}